\documentclass[lettersize,journal]{IEEEtran}
\usepackage{cite}
\usepackage{amsmath,amssymb,amsfonts}
\usepackage{graphicx}
\usepackage{svg}
\usepackage{textcomp}
\usepackage{xcolor}
\usepackage{algorithm}
\usepackage{algpseudocode}
\usepackage{glossaries}
\usepackage{subcaption}
\usepackage{float}
\usepackage{svg}
\usepackage{hyperref}

\begin{document}

\title{An Informative Planning Framework for Target Tracking and \\ Active Mapping in Dynamic Environments with ASVs}

\author{Sanjeev Ramkumar Sudha, Marija Popovi{\'c}, and Erlend M. Coates
\thanks{Sanjeev R.S. and Erlend M. Coates are with the Norwegian University of Science and Technology (NTNU), Ålesund, Norway. {\tt\small\{sanjeev.k.r.sudha, erlend.coates\}@ntnu.no}. Marija Popovi{\'c} is with the MAVLab, Faculty of Aerospace Engineering, TU Delft, The Netherlands. {\tt\small m.popovic@tudelft.nl}.}
}



\maketitle

\begin{abstract}
Mobile robot platforms are increasingly being used to automate information gathering tasks such as environmental monitoring. Efficient target tracking in dynamic environments is critical for applications such as search and rescue and pollutant cleanups. In this letter, we study active mapping of floating targets that drift due to environmental disturbances such as wind and currents.
This is a challenging problem as it involves predicting both spatial and temporal variations in the map due to changing conditions. 
We introduce an integrated framework combining dynamic occupancy grid mapping and an informative planning approach to actively map and track freely drifting targets with an autonomous surface vehicle.
A key component of our adaptive planning approach is a spatiotemporal prediction network that predicts target position distributions over time. We further propose a planning objective for target tracking that leverages these predictions. Simulation experiments show that this planning objective improves target tracking performance compared to existing methods that consider only entropy reduction as the planning objective. Finally, we validate our approach in field tests, showcasing its ability to track targets in real-world monitoring scenarios.
\end{abstract}


\section{Introduction}
\IEEEPARstart{T}{arget} search and monitoring with autonomous mobile robots has many applications, such as search and rescue \cite{meera2019obstacle, ai2019intelligent}, cleaning pollutants \cite{barrionuevo2025optimizing}, and wildlife monitoring \cite{tokekar2013tracking}. Mobile robot platforms such as autonomous surface vehicles (ASVs), and uncrewed aerial vehicles (UAVs) allow for faster and more efficient search \cite{dunbabin2012robots} compared to traditional approaches such as manned or teleoperated surveys \cite{murphy2017disaster} and static sensor networks \cite{lanzolla2021wireless}. For target tracking in \textit{a priori} unknown dynamic environments, a robot must trade off between exploring previously unseen regions to discover new targets and acquiring new observations of previously observed moving targets to update their position estimates. Monitoring in such dynamic environments remains a challenging problem.

In this work, we consider the tracking of freely drifting targets with an ASV. We aim to build and maintain an accurate map of target positions in a dynamic environment for any number of targets with unknown initial positions. Partially submerged objects are subject to external disturbances such as wind and currents in the marine environment. This is a pertinent issue for maritime search and rescue \cite{ai2019intelligent}, cleanups of plastic waste \cite{barrionuevo2025optimizing}, or other pollutants \cite{wang2019dynamic}. 
\begin{figure}[!htbp]
    \centering
    \includegraphics[width=0.82\linewidth]{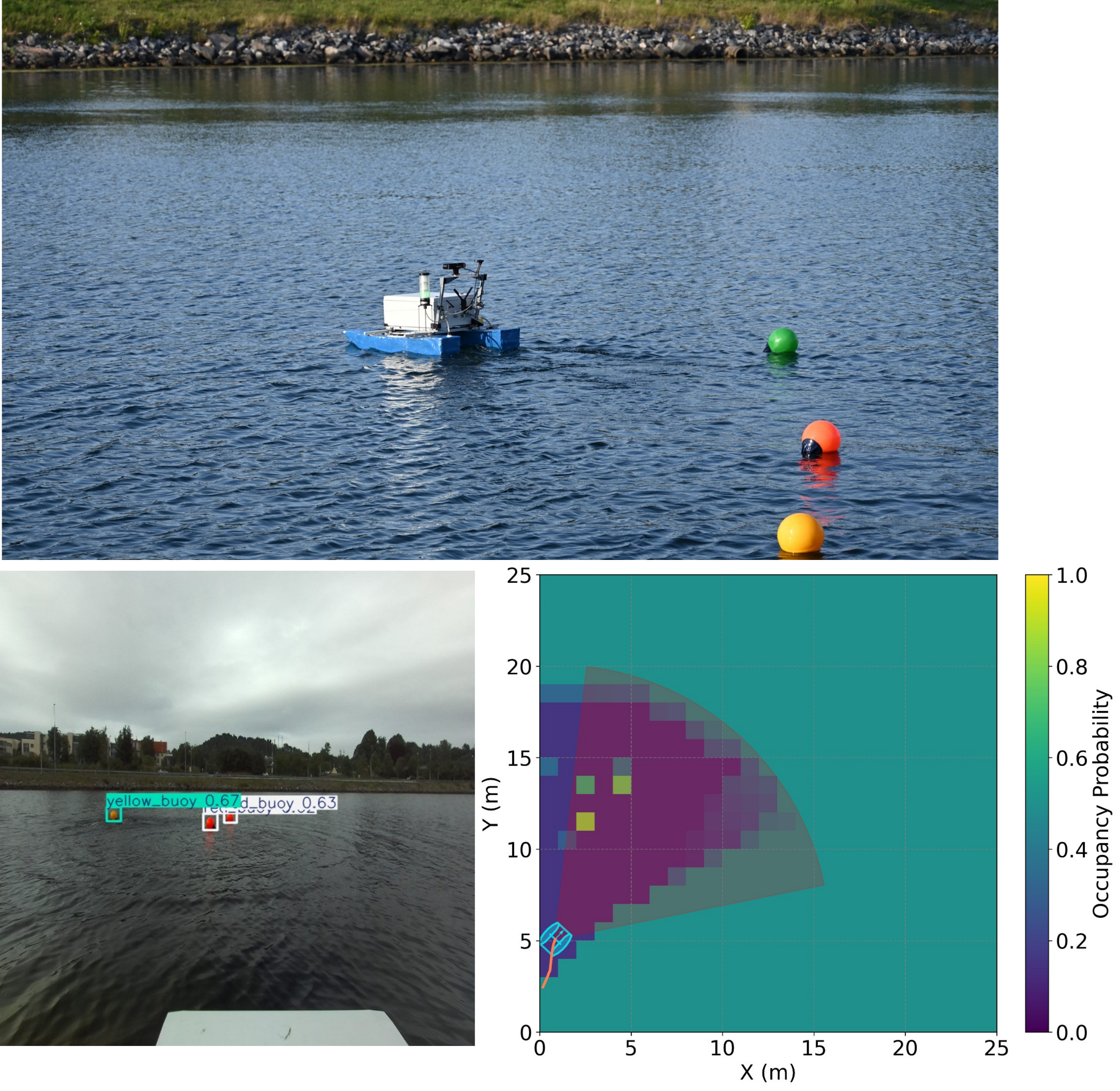}
     \caption{Our informative path planning (IPP) framework applied to monitoring freely drifting targets (colored buoy markers) with an autonomous surface vehicle that is equipped with a stereo camera. A dynamic mapping strategy is used for target tracking. We introduce a spatiotemporal prediction network and a new IPP objective that enable accurate target tracking.
     }
    \label{fig:experiment_1}
\end{figure}

Methods for finding sampling paths that maximize the information gathered for resource-constrained robots are termed \textit{informative path planning} (IPP) \cite{popovic2024learning}. Informative trajectories are quantified by an information-theoretic measure representing the mission objective, such as entropy in the map. For static environments, entropy is generally used as the planning objective for efficient spatial coverage \cite{meera2019obstacle, popovic2020informative}. In dynamic environments, predicting the spatial and temporal variations in the map and frequent replanning are required to maintain an accurate map of target positions.

Some studies use numerical models of known physical processes to map the environment \cite{wang2019dynamic, du2021multi}. In contrast, we focus on tracking discrete targets that drift, which are difficult to model due to the stochastic nature of environmental disturbances \cite{breivik2011wind}. Classical target tracking methods detect, associate identifiers with targets, and track them individually \cite{robin2016multi}. 
Thus, the computational demand also scales with the number of targets. 
Using occupancy grid mapping for target tracking \cite{coue2006bayesian} provides a simpler alternative by handling occupancy probabilities of the targets in a global probabilistic map representation, avoiding maintaining individual tracks and associating identifiers to targets. Prior works \cite{wolf2004online, meyer2012occupancy, coue2006bayesian} on occupancy grid mapping in dynamic environments learn state transitions online, but such methods can become unsuitable for real-time applications due to computational constraints.

To overcome these limitations, we present an integrated solution for actively mapping moving targets in dynamic environments. A dynamic occupancy grid mapping approach is used to update the map, combining the sensor observations and an approximation of the targets' drift using the measured wind velocity at each time step. We employ adaptive sampling-based planning to enable the robot to track targets by driving it towards regions of interest. We introduce a spatiotemporal prediction network to estimate uncertain target positions for any number of targets. This enables efficient predictions for longer time horizons, based only on the current map state and the measured wind speed. Furthermore, our proposed planning objective explicitly considers target tracking uncertainty by using target position predictions from the network.

We make the following claims supported by extensive simulation experiments and field trials of our IPP framework implemented on an ASV: (i) Our proposed IPP objective that utilizes the spatiotemporal network improves target tracking performance when compared to existing methods that use only entropy reduction as the planning objective, and (ii) incorporating drift predictions in the mapping stage improves mapping accuracy when compared to mapping without predictions. 
As a result, our framework enables efficient real-time adaptive planning and active mapping in dynamic environments, as also validated through field trials. We open-source our software at \url{https://github.com/sanjeevrs2000/ipp_dyntrack}.

\section{Related Work}
\label{sec:rw}

Various studies have investigated IPP approaches for mapping discrete and static targets \cite{meera2019obstacle, vashisth2024deep, moon2022tigris}. Barrionuevo et al. \cite{barrionuevo2025optimizing} use reinforcement learning for planning to monitor and collect floating waste with a fleet of two ASVs, assuming perfect localization of the targets. Chen and Liu \cite{chen2019multi} study IPP for monitoring in dynamic environments with a model-predictive tree search with multi-objective optimization. They, however, assume the environmental dynamics to be deterministic when in reality, long-term predictions are difficult due to the stochastic nature of real-life environments. Similar to \cite{chen2019multi}, we use a sampling-based planning approach for IPP that incorporates both the spatial and temporal variations in the map in the planning objective. In addition, our proposed spatiotemporal prediction network informs the informative planner of the uncertainties in target positions over the planning horizon, taking into account the stochastic nature of the wind and currents.

Some literature in robot monitoring for source seeking and plume tracking uses high-fidelity numerical models of the underlying physical processes \cite{preston2024phortex, wang2019dynamic, du2021multi, ma2025adaptive}. However, their adaptive planning strategies are often limited to myopic behaviors, i.e., single-step lookahead, due to computational and operational constraints. A body of existing work also addresses robot monitoring in dynamic environments with multiple robots \cite{du2021multi, wang2019dynamic}. Dividing roles between a team of robots simplifies decision making when there are multiple objectives \cite{wang2019dynamic, barrionuevo2025optimizing}. In contrast, we focus on tracking drifting targets with a single ASV while noting that most of our framework could still be applicable in a multirobot setting. 

The occupancy grid mapping algorithm \cite{elfes1989using} is widely used in robotics to create probabilistic occupancy maps for static environments. Several studies extend the occupancy grid formulation for mapping in dynamic environments, as relevant to our problem setup. Meyer et al. \cite{meyer2012occupancy} propose a framework for mapping in dynamic environments where the state transition probabilities are learned. Wolf and Sukhatme \cite{wolf2004online} study mapping in dynamic environments by maintaining separate maps for static and dynamic parts. The Bayesian occupancy filter \cite{coue2006bayesian} introduces a dynamic occupancy grid mapping method for multi-object tracking. The map is updated in two steps, namely the estimation and prediction steps, similar to Bayesian state estimation. While our method is similar to that of Coue et al. \cite{coue2006bayesian}, we infer the measured wind speed for prediction at each time step rather than learning it through observations. It allows us to maintain a map that is agnostic to the number of targets while retaining the probabilistic formulation. This dynamic occupancy grid, combined with predictions from the spatiotemporal network for efficient replanning, provides a framework capable of real-time mapping and planning. 

Various recent studies apply deep learning-based methods for mapping temporally varying environments. Recurrent neural networks have been used to predict future occupancy states in dynamic environments in autonomous driving scenarios \cite{mann2022predicting,toyungyernsub2021double}. Such methods rely on continuous observations of the targets of interest and predict the future dynamic states in the map for only a few seconds. 
Wang et al. \cite{wang2023spatio} employ an attention-based neural network for persistent monitoring of mobile targets. In contrast to these works, we propose a physics-informed spatiotemporal prediction model that is independent of the number of targets and efficiently estimates uncertain target position distributions, for longer horizons of up to $30$ s. Overall, our approach enables efficient tracking of moving targets with unknown initial positions in dynamic environments.
This enables effective autonomous monitoring for applications such as search and rescue and pollutant cleanups.

\section{Problem Statement}

Our aim is to detect and actively map targets of interest, such as floating waste, influenced by environmental disturbances in an \textit{a priori} unknown environment. We frame this as an IPP problem \cite{popovic2024learning}. We seek a trajectory $\mathcal{T}$ from the set of all feasible trajectories $\Psi$, that maximizes an information-theoretic measure or utility $I(\mathcal{T})$, given a limited budget ${B}$, such as mission time. In particular, the goal is to approximate the optimal trajectory $\mathcal{T}^*$ given by
\begin{equation}
\mathcal{T}^*= \underset{\mathcal{T} \in \Psi}{\arg\max} \, I(\mathcal{T}) \text { s.t. } C(\mathcal{T}) \leq \ {B}\, ,
\label{eq:ipp}
\end{equation}
where $C(\mathcal{T})$ is the cost associated with the trajectory $\mathcal{T}$. 
In this work, we consider constant speed trajectories. The trajectory planning problem thus reduces to searching for a geometric path that maximizes $I(\mathcal{T})$.

\section{Our Approach}
\label{sec:approach}

\subsection{Overview}
\begin{figure}[!h]
    \centering
    \includegraphics[width=0.95\linewidth]{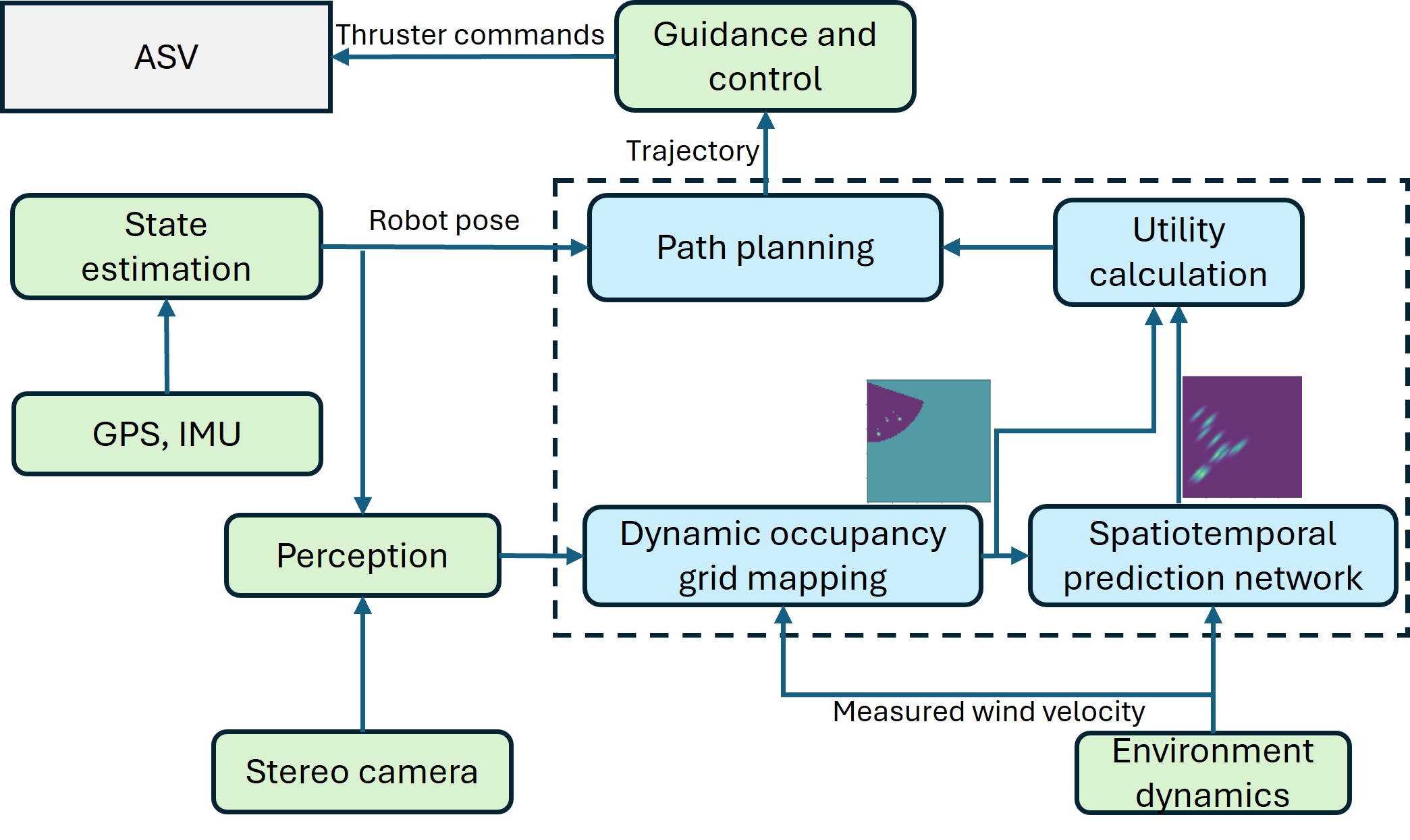}
    \caption{An overview of our IPP framework (in blue) for active mapping of dynamic targets. We detect and localize targets with a stereo camera on an ASV. We use a dynamic occupancy grid mapping method that updates the map with both the sensor percepts and predicted motion of the targets. We introduce our spatiotemporal network that enables efficient predictions to aid the adaptive planning approach.
    }
    \label{fig:overview}
\end{figure}

We introduce a novel IPP framework for actively mapping moving targets in dynamic environments. An overview of this framework is shown in \autoref{fig:overview}. We consider an ASV with a stereo camera for object detection and localization. A probabilistic occupancy grid is used to map these detections. We also incorporate a prediction step in the mapping to compensate for the targets' drift. A key aspect of our approach is a spatiotemporal prediction network to estimate the spatial and temporal variations of the target positions over future time steps, given the current state of the occupancy grid and the current wind conditions. We then propose a new IPP utility formulation for target tracking that leverages predictions from our spatiotemporal network.

\subsection{Perception}
\label{sec:sm}

The ASV is equipped with a forward-facing stereo camera to detect and localize targets of interest. In this work, we use YOLOv8 \cite{yolov8_ultralytics} to identify the targets of interest, noting that our approach is agnostic to the choice of object detection algorithm. The positions of detected targets are projected onto the global occupancy map: 
\begin{equation}
\boldsymbol{p}_{g} =
\underbrace{
\begin{bmatrix}
c & -s & 0 \\
s & c & 0 \\
0 & 0 & 1
\end{bmatrix}
}_{R_{z\psi}}
\underbrace{
\begin{bmatrix}
0 & 0 & 1\\
-1 & 0 & 0\\
0 & -1 & 0
\end{bmatrix}
}_{R_{bc}}
\boldsymbol{p}_{{c}}
+
\begin{bmatrix}
x_v \\
y_v \\
0
\end{bmatrix} \, ,
\label{eq:frame_transf}
\end{equation}
where $\boldsymbol{p}_{c}$ is the estimated depth of the target in the camera frame, $\boldsymbol{p}_{g}$ is its position in the global east-north-up (ENU) frame. The rotation matrix for transformation from the camera to the body frame is denoted by $R_{bc}$, assuming the camera is forward-facing and the distance between the body's origin and the camera is negligible. The body to global frame rotation matrix is given by $R_{z\psi}$, where $c$ and $s$ denote $\cos\psi$ and $\sin\psi$. The estimated global position and heading of the ASV are denoted by $(x_v, y_v)$ and $\psi$, respectively.


\begin{figure}[!htbp]
\centering
    \begin{subfigure}{0.24\textwidth}
    \centering
    \includegraphics[width=0.97\linewidth]{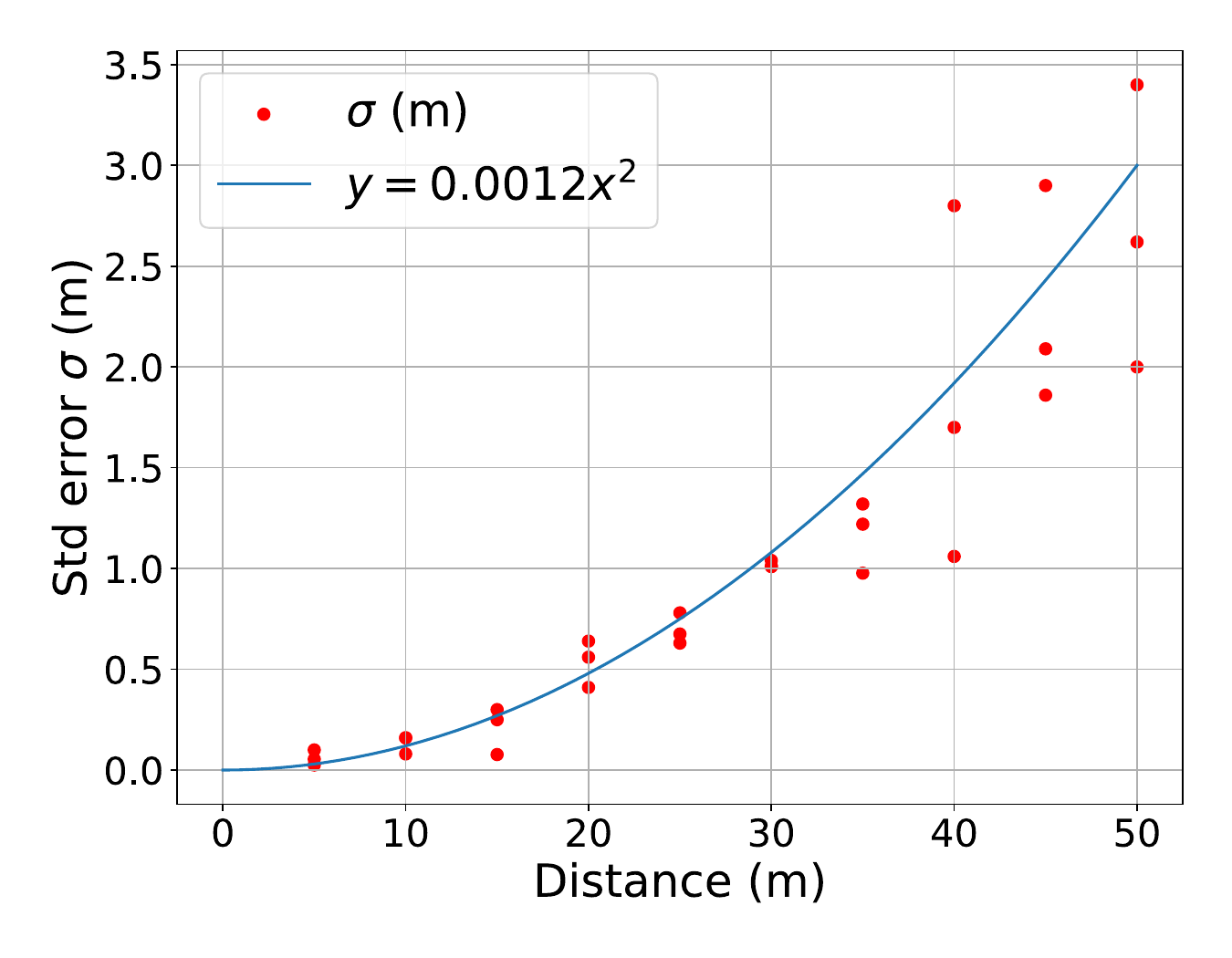}
\end{subfigure}
\begin{subfigure}{0.24\textwidth}
    \centering
    \includegraphics[width=0.97\linewidth]{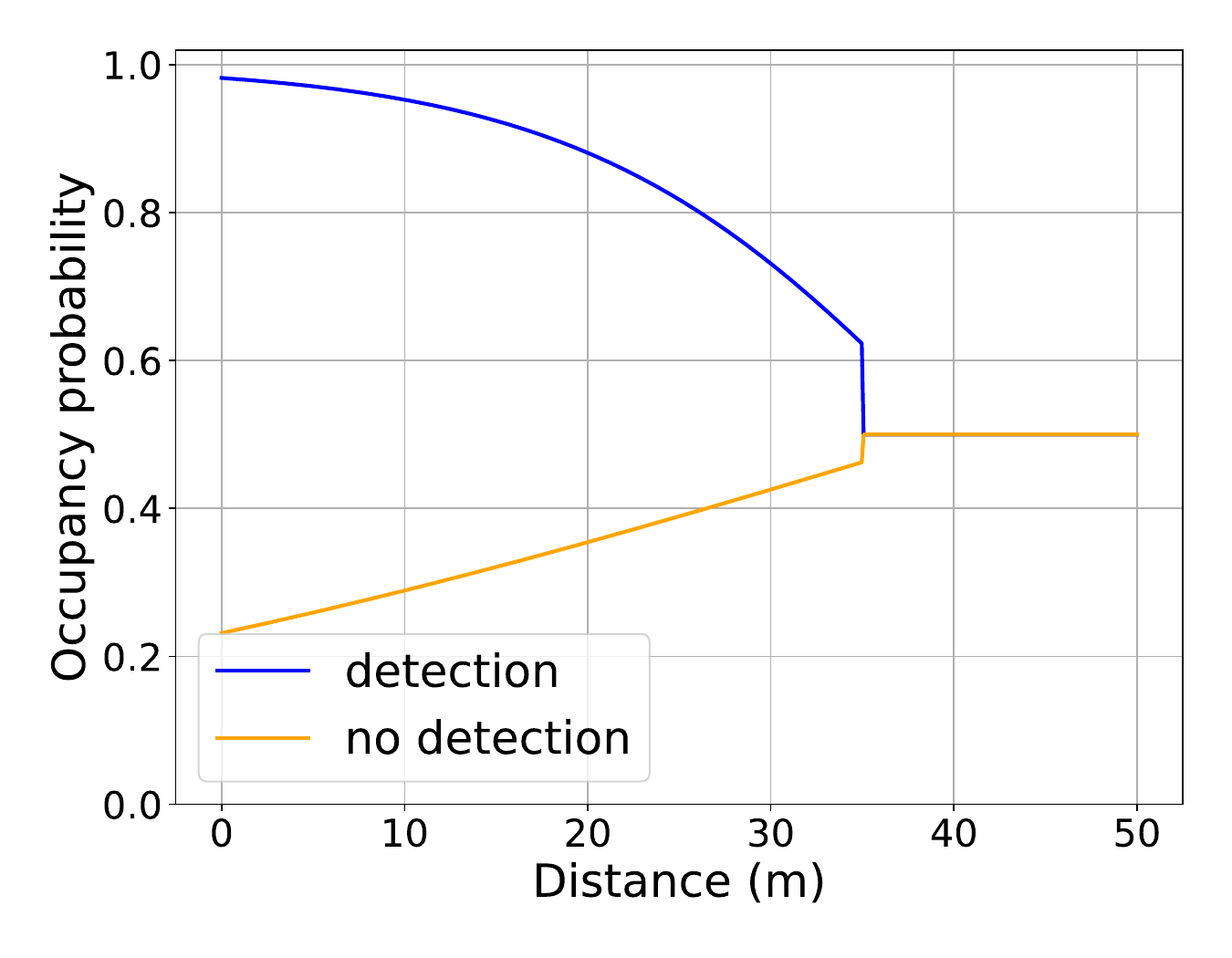}
    \end{subfigure}
\caption{Localization error $\sigma(r)$ as function of distance to detection $r$ based on empirical analysis (left). The inverse sensor models used for mapping are shown on the right.}
\label{fig:perception}
\end{figure}
Occupancy grid mapping requires inverse sensor models to convert sensor observations to belief probabilities on the map. To define inverse sensor models for our application, we perform an empirical analysis in simulations to estimate the localization error of the detected objects. The inverse sensor models we use are similar to previous work \cite{popovic2020informative, moon2022tigris}, but are kept asymmetric to better suit our problem setup in dynamic environments. Another distinction is that the inverse sensor model for detection also associates an occupancy probability for neighboring cells of each detection with a Gaussian kernel to take into account the localization errors in the detections. We derive the standard deviation $\sigma(r)$ of this error as a function of the distance to the detection $r$, through empirical analysis. To do this, we place a single object at various distances and relative heading angles to the camera in the simulation environment. We show the resulting standard deviations in the errors of the estimated positions of the targets in \autoref{fig:perception}. A curve is fit over this plot to estimate error as a function of the distance. Our inverse sensor models, for detections ($z$), and free cells ($z'$), are defined as:
\begin{align}
p\!\left(m_{i,j} \mid z\right)
&= \frac{1}{1+e^{a(r-d)}} \cdot
   e^{-\tfrac{\lVert \boldsymbol{x}_c - \boldsymbol{x}_d \rVert^2}{2\sigma^2}},
\label{eq:sensor_model}\\
p\!\left(m_{i,j} \mid z'\right)
&= \frac{1}{1+e^{a'(r-d')}} \, ,
\label{eq:negative_sensor_model}
\end{align}

\noindent 
where the standard deviation $\sigma(r)$ is set to  $0.0012r^2$, based on the analysis in \autoref{fig:perception}. The estimated position of the detected target is given by $\boldsymbol{x}_d$, and $||\boldsymbol{x}_c - \boldsymbol{x}_d||$ denotes the distance of a nearby cell $c$ to this detection. The inverse sensor model associating a detection probability to cell $(i,j)$ given a detection is denoted by $p(m_{i,j}|z)$, parameterized by $a, d$. The negative inverse sensor model for updating free cells is given by $p(m_{i,j}|z')$, parameterized by $a', d'$. The radial field of view (FOV) is clipped at $35$ m, as depth measurements become unreliable for further detections. The inverse observation models are depicted in \autoref{fig:perception}. 
%

\subsection{Dynamic Occupancy Grid Mapping}
\label{subsec:map}
We introduce a dynamic occupancy grid method for mapping to keep track of detected target positions. The map is updated in two stages at each time step, namely the estimation and prediction steps. First, in the estimation step, the map is updated with the sensor observations, using the classical occupancy grid algorithm \cite{elfes1989using}. In the prediction step, we compensate for the drift of targets in the map with the instantaneous wind speed. In the estimation step, we update the map recursively as: 
\begin{align}
l(m_{i,j} | z_{1:t}) = l(m_{i,j} | z_{1:t-1}) + l(m_{i,j}|z_{t}) -l(m_{i,j}) \,,
\label{eq:og_update}
\end{align}
where $z_t$ is the sensor observation at time step $t$. The log odds notation, $l(.) = \log \big( p(.) / (1 - p(.))\big)$, of the occupancy probability of cell $(i,j)$, $p(m_{i,j})$, is denoted by $l(m_{i,j})$.
The log odds representation of the inverse sensor model $p(m_{i,j}|z_{t})$ is denoted by $l(m_{i,j}|z_{t})$.

In the prediction step, the occupancy grid is updated each time step to account for the drift of the targets. 
We assume that the drift velocity of small floating objects, such as the ones we consider, can be approximated to be proportional to the instantaneous wind velocity, in line with previous work \cite{rohrs2015drift}. 
The drift for a single time step can therefore be approximated as:
\begin{equation}
\begin{gathered}
dx =  {\gamma} \, v_{w}\cos({\psi_{w}}) \, dt  \, ,  \\
dy =  {\gamma} \, v_{w}\sin({\psi_{w}}) \, dt      \, ,
\label{eq:drift}
\end{gathered}
\end{equation}
where $v_{w}$ is the wind speed, $\psi_{w}$ is the direction of the wind velocity at time $t$, and $dt$ is the time step between successive updates of the map. The wind factor $\gamma$ is defined as the ratio of the drift speed of the object to the wind speed, which we set to be $0.03$ following \cite{rohrs2015drift}. To update the occupancy grid with the drift prediction of the targets, we first filter it in order to keep only occupied cells ($>0.5$) as:
\begin{equation}
    m'_{i,j} \;=\; 
        \begin{cases}
          m_{i,j}, & m_{i,j}>0.5 \, ,\\
          p_{\mathrm{low}}, & \text{otherwise} \, ,
        \end{cases}
\label{eq:dynamic_og_filter}    
\end{equation}
where $m'_{i,j}$ denotes the occupancy probabilities in the filtered grid $\mathcal{M'}$. As the incremental drift at each time step is much smaller than the resolution of the occupancy grid, we store the accumulated drift in variables $R_{x}, R_{y}$. In the prediction step, the occupancy probabilities are updated only if $\texttt{round}(R_x)>0$ or $\texttt{round}(R_y)>0$. The update rule for the prediction step is described in lines \autoref{line:dyn_update1}-\autoref{line:dyn_update2} of Algorithm \autoref{alg:mapping}, where $\alpha$ and $\beta$ are parameters to trade off between the current occupancy state and the prediction. The occupancy probabilities in the grid are constrained to the interval $[p_{\mathrm{low}}, p_{\mathrm{high}}] =[0.15,\,0.9]$ to adapt the mapping to the dynamic setting. A condensed form of this dynamic occupancy grid mapping algorithm is given in Algorithm \autoref{alg:mapping}.


\begin{algorithm}[!htbp]
  \caption{Dynamic occupancy grid mapping}
  \label{alg:mapping}
\begin{algorithmic}[1]
  \Ensure Occupancy grid $\mathcal{M}$
  \State Initialize $R_{x} \gets 0,\; R_{y}\gets0$, $\mathcal{M} \;=\; {0.5}$
  
  \For{$t = 0 : {B}$}
    \State $O\gets\phi$   
    
    \For{each detection $d$}       \Comment{Estimation step}
      \State Localize $d$ at $\boldsymbol{x}_d$
      \State Find cells $c$ around $d$ s.t. \eqref{eq:sensor_model} $ > p_{\mathrm{low}}$
      \State Update $m_{i,j}$ for all $c$ with \eqref{eq:sensor_model}
      \State $O\gets O\cup c$
    \EndFor
    \State $F\gets\mathrm{fov}(\boldsymbol{x}^{t})\setminus O$
    \For{each $(i,j)\in F$}
      \State Update with \eqref{eq:negative_sensor_model}
    \EndFor

    \State Get $\mathcal M'$ according to \eqref{eq:dynamic_og_filter}      \Comment{Prediction step}

\State Compute drift $(dx, dy)$ according to \eqref{eq:drift}
\State $R_{x} \gets R_x + dx / \Delta_{x}$, $R_y \gets R_y + dy / \Delta_{y}$
\State $s_x \gets \texttt{round}(R_x)$, $s_y \gets \texttt{round}(R_y)$
\State $R_x \gets R_x - s_x$, $R_y \gets R_y - s_y$
\ForAll{cells $(i,j)\in\mathcal M$} \label{line:dyn_update1}
      \If{$m'_{i,j}\neq m'_{i-s_x,j-s_y}$}
        \State $m_{i,j}\gets\alpha\,m'_{i-s_x,j-s_y}+\beta\,m_{i,j}$ \label{line:dyn_update2}
      \EndIf
\EndFor
  \EndFor
\end{algorithmic}
\end{algorithm}

\subsection{Spatiotemporal Prediction Network}

Maintaining an accurate map of target positions requires planning informative trajectories for the robot that facilitate repeated detections of targets. This involves predicting changes in the map over some time frame. We introduce a spatiotemporal prediction network to predict the uncertainty of the target positions in the map over a certain time horizon. Although our dynamic occupancy mapping method contains a prediction step to estimate the drift of targets, it is not accurate at predicting drift for longer time horizons, due to the stochastic nature of environmental disturbances. Therefore, we create a formulation to estimate the spatial and temporal variations of the target positions for future time steps.

We assume that the position of each target is represented by a Gaussian kernel, with the uncertainty in its position estimate growing with time, similar to previous work \cite{wang2023spatio}. To reflect this assumption, we assume that the variances of the kernel are proportional to the wind speed and increase with the prediction time, with the principal axis being along the direction of the measured wind. We assume that the centers of each kernel translate by distances calculated as per \eqref{eq:drift}. The variance along the direction of the wind is assumed to be greater, as for most realistic scenarios, the wind speed varies faster than the direction \cite{kaimal1994atmospheric}. 
Based on this assumption, the standard deviations of the kernel along the major and minor axes, $\sigma_{\parallel}$, and $\sigma_{\perp}$ for a prediction interval $t$ are set as $\sigma_{\parallel}=0.5\,\gamma v_{w}t$, and $\sigma_{\perp}=0.2\,\gamma v_{w}t$ respectively. In case of negligible wind speeds ($v_w \approx0$), $\sigma_{\parallel}=\sigma_{\perp}=0.1\,t$.

\begin{figure}[!htbp]
    \centering
    \includegraphics[width=0.96\linewidth]{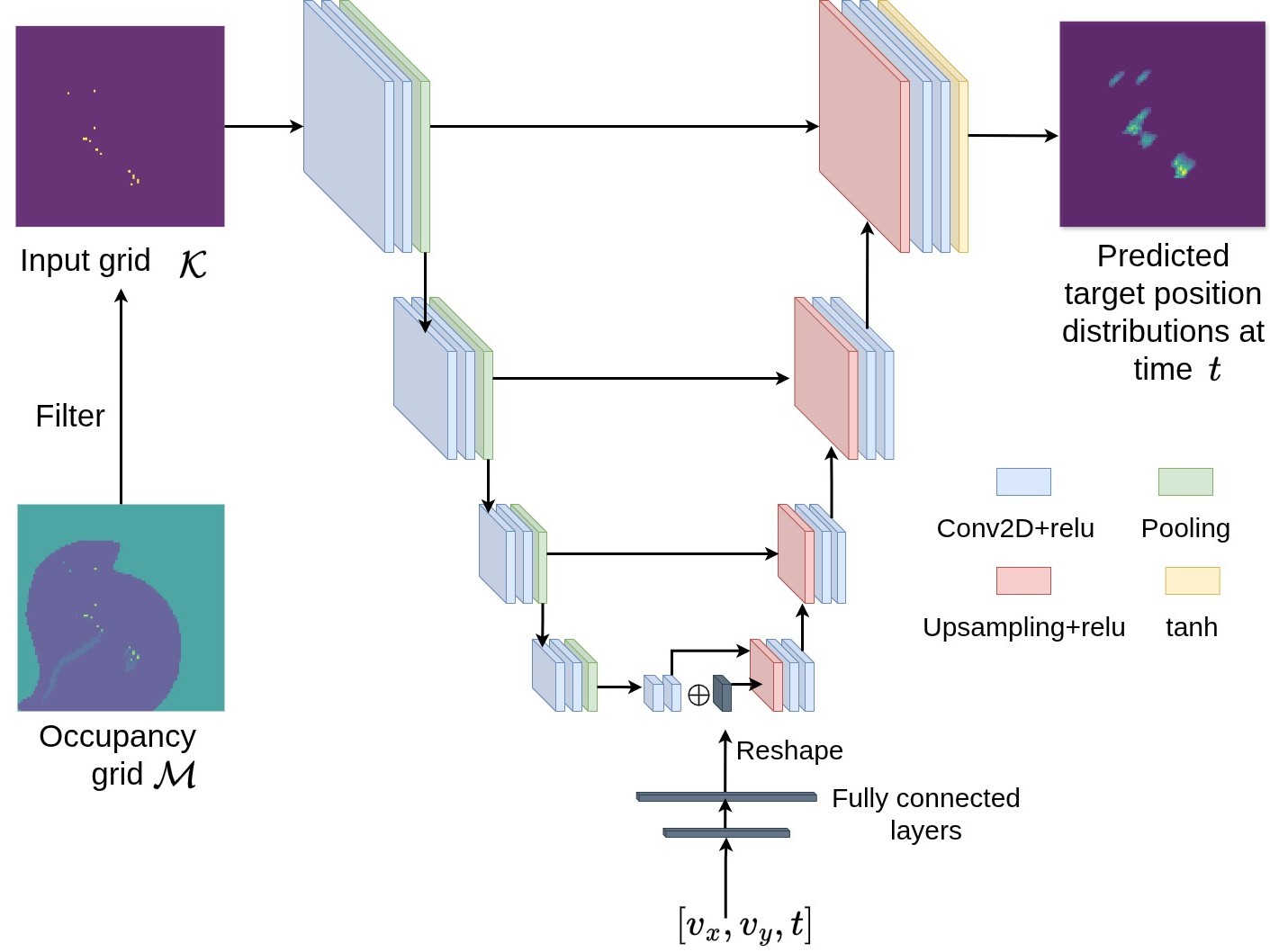}
    \caption{An overview of our spatiotemporal prediction network. The inputs are a filtered grid with only target positions, along with a vector of the wind speeds $v_x$, $v_y$, and prediction interval $t$. The output is the predicted target position distributions at $t$.}
    \label{fig:nn}
\end{figure}

We train our spatiotemporal prediction network in a supervised fashion to obtain these predictions. The network enables us to get these predictions for multiple prediction steps for any number of targets in an efficient way. An illustration of this spatiotemporal prediction network is shown in \autoref{fig:nn}. The input to the network is a filtered binary grid $\mathcal{K}$ with the target positions, which is obtained as follows:
\begin{equation}
    k_{i,j} \;=\; 
        \begin{cases}
          1, & m_{i,j}>0.5 \,  ,\\
          0, & \text{otherwise} \, .
        \end{cases}
\label{eq:filter_grid_prediction}    
\end{equation}

We provide the wind parameters $v_w$, $\psi_w$, and the prediction interval as other inputs to the network. The network output is a grid of the same size with predicted position uncertainties of the targets at an arbitrary time step. We use a UNet architecture \cite{ronneberger2015u}, which is widely used for image-to-image prediction tasks such as segmentation. The wind parameters and the prediction time are embedded into the latent space of the network. 

To train our network, a synthetic dataset is created to simulate input conditions with different wind conditions, target positions, and prediction intervals. To construct inputs resembling target occupancies in the map, binary grids of size $100 \times 100$ containing an arbitrary number of objects ($<20$) with random positions are generated. For each of these input grids representing the spatial distribution of the targets, outputs are computed as described above, for various combinations of wind parameters and prediction time steps. 
While predicting the drift of objects on the water surface is a difficult problem \cite{breivik2011wind}, we find that these assumptions suit our problem setup. Using an input representation that is agnostic to the number of targets makes our method generalizable, as validated through both simulation and field experiments, in Secs. \ref{sec:results} and \ref{sec:field_exp}.



\subsection{Informative Planning for Dynamic Target Tracking}
\label{subsec:planning}

The robot needs to replan at frequent intervals to account for the changing environment. It must explore unvisited regions of the map while redetecting targets. Planning informative trajectories in dynamic environments necessitates predicting both spatial and temporal variations in the environment. Commonly used IPP objectives are based on Shannon entropy measured based on the map state \cite{popovic2020informative, moon2022tigris}. However, these formulations do not explicitly account for temporal variations and periodic target redetections. In contrast, we exploit the predictions from our spatiotemporal network to include the temporal uncertainties in target positions in the planning objective.

To trade off between exploring the environment and redetecting targets, we introduce a new planning utility that consists of both exploration and target tracking objectives. We propose a new target tracking component in the utility that explicitly considers possible target detections by estimating the spatiotemporal variations in the targets' position estimates in the planning objective. It is computed with the predictions from our spatiotemporal network, for prediction steps over the considered planning horizon. To compute the predicted change in entropy, the dynamic occupancy grid mapping is forward-simulated for predicted measurements along considered trajectories. We define the planning utility as:
\begin{align}
I(\mathcal{T}) &= \Big[ H(\mathcal{M}) - H(\mathcal{M} \,|\, z_{\mathcal{T}}) \Big] \space + \frac{w}{|\mathcal{T}|}  \sum_{t}{J(\boldsymbol{x}^{t})} \, ,
\label{eq:reward} \\
J\left(\boldsymbol{x}^{t}\right) &=  \sum_{i, j} \frac{e^{-2\cdot p_{ij}(t)}}{|\texttt{fov}(\boldsymbol{x}^{t})|}, \,\forall (i,j)\in \texttt{fov}(\boldsymbol{x}^{t}) \, ,
\label{eq:track}
\end{align}
where $H(\mathcal{M})$ denotes the mean Shannon entropy of the occupancy grid, and $H(\mathcal{M}|z_{\mathcal{T}})$ denotes the predicted posterior mean entropy after simulating measurements $z_{\mathcal{T}}$ along a trajectory $\mathcal{T} = \{\boldsymbol{x}^t | t = 1,\dots,|\mathcal{T}|\}$. The target tracking objective is defined as the averaged sum of the utility $J(\boldsymbol{x}^{t})$ over the prediction steps, where $\boldsymbol{x}^{t}$ is a pose on the trajectory corresponding to time $t$, and $p_{ij}(t)$ is the occupancy probability of cell $(i,j)$ as output by the network. This explicitly considers tracking by rewarding possible detections with uncertain positions over the planning horizon. The weighting factor $w$ determines the importance given to the two objectives, i.e., exploration and target tracking. In contrast to existing work \cite{meera2019obstacle, moon2022tigris}, our proposed target tracking utility promotes efficient target tracking by considering possible detections along the paths with the associated uncertain spatiotemporal variations in the target positions. 

For planning, we use a sampling-based planner to leverage these lookahead predictions. We consider a set of candidate paths $\Psi$ constructed using a finite set of evenly discretized heading changes, assuming a constant speed for the ASV. These paths are smoothed with Dubins curves to generate dynamically feasible paths for the ASV. The planning objective as described in \eqref{eq:reward}-\eqref{eq:track}, is calculated for each of these candidate paths $\mathcal{T}$.

\section{Results and Discussion}
\label{sec:results}

\subsection{Experimental Setup}

We perform simulations with an ASV in Gazebo with the VRX simulator \cite{bingham2019toward}. The wind is simulated as a stochastic process with the slowly-varying component modeled as a first-order Gauss-Markov process. We consider environments of size $100$\,m\,$\times\,100$\,m discretized with size $\Delta_x=1$\,m, $\Delta_y=1$\,m. We generate spherical buoy markers in the simulations as targets of interest and model the wind forces on these targets as described by Fossen \cite{fossen2011handbook}. Different scenarios are generated by changing the mean wind speed, mean direction, and the spawn positions of the targets. Eight targets are present in the environment, and the mean wind speed is varied between $6$-$10$\,m/s in the scenarios. We run Monte Carlo simulations under these environmental conditions to perform a statistical study of the different components of our framework. The mapping is performed at $5$ Hz. The planning horizon for replanning is $25$ s, with time steps of $1$ s. The replanning step for seven candidate paths takes  $\approx 0.08$ s, on a computer equipped with an i7-12700H processor, 32GB DDR5 RAM and RTX3050 GPU, for parallelized computations of the utility \eqref{eq:reward}. Each Monte Carlo trial is run for a mission time of $B=250$ s, and the ASV's speed is set to $1.5$\,m/s. 
For path-following control, line of sight guidance \cite{fossen2011handbook} is used along with PID speed and heading controllers.

We consider the following metrics for comparison: the reduction in the mean Shannon entropy in the occupancy map ($\Delta H$), the mean squared error (MSE), and the mean detections ($\overline{N}$). The mean number of detections is computed as the average of the total number of detections, i.e., sum of the detections at each time step. The MSE between the ground truth and the occupancy grid is computed as \cite{barrionuevo2025optimizing}:
\begin{equation}
    \frac{\sum  \bigl( \mathcal{M} - G \circledast \mathcal{Y} \bigr)^2}{m\times n} \, ,
\label{eq:mse}
\end{equation}
\noindent where $G$ is a Gaussian filter, $\mathcal{Y}$ is the ground truth from the simulation environment, $\circledast$ is the convolution operator, $m$, $n$ denote the height and width of the occupancy grid. The error is computed between the occupancy grid $\mathcal{M}$ and $G \circledast \mathcal{Y}$ as $\mathcal{Y}$ is a sparse matrix.


\subsection{Ablation Study - Planning Utility}

To study the effect of our spatiotemporal prediction network and the proposed planning utility, we compare various values of the weighting factor $w$ in \eqref{eq:reward}, which determines the trade-off between the two objectives: reducing entropy in the occupancy grid and redetecting targets. Monte Carlo simulations are performed for $w=0,\, 2,\,5,\,5\,t/B$, and $5(1-t/B)$, with the sampling-based planner executed in a receding horizon fashion. The resulting metrics are shown in \autoref{tab:table2}. Higher values of $w$ correspond to the planner prioritizing target redetection, while $w=0$ corresponds to an exploratory planner that only considers the entropy reduction in the occupancy grid in the utility. 


\begin{table}[!htb]
    \centering
\centering
\begin{tabular}{|c|c|c|c|}
\hline

${w}$ & ${\Delta H}$ & ${\overline{N}}$ & {MSE}\\
\hline
$0$ & $\mathbf{0.345 \pm 0.012}$ & $1.122 \pm 0.307$  & $\mathbf{0.046 \pm 0.005}$\\
$2$ & $0.314 \pm 0.019$ & $	1.594 \pm 0.193$ & $0.065 \pm 0.011$ \\
$5$ & $0.295 \pm 0.019$ & $1.690 \pm 0.142$ & $	0.076 \pm 0.011$\\
$5\, t/B$ &$ 0.287 \pm 0.014$& $1.670 \pm 0.131$ & $	0.081 \pm 0.008$\\
$5(1 - t/B)$ & $0.293 \pm 0.021$ & $\mathbf{1.696 \pm 0.137}$ & $0.077 \pm 0.012$\\
\hline
\end{tabular}
    \caption{Comparative analysis for different values of $w$ with the best values for each metric highlighted in bold. The trials with $w=5(1-t/B)$ are the best at target tracking, while the exploratory planner ($w=0$) is the least effective.}
    \label{tab:table2}\end{table}

\autoref{tab:table2} reveals that the purely exploratory case ($w=0$) is the best at reducing entropy and subsequently MSE. It is also worse at detecting targets by at least $\approx 31\%$ as compared to the other variants that use the tracking information gain in the planning objective. This highlights the importance of our prediction network and the corresponding target tracking utility in the adaptive planning strategy to re-identify targets. 
Further, we observe that higher values of $w$ correspond to more target detections. These results demonstrate that our IPP utility improves target tracking performance by enabling more frequent target reacquisition.

\subsection{Ablation Study - Mapping}

\begin{figure}[!htb]
    \centering
    \begin{subfigure}{0.23\textwidth}
        \centering
        \includegraphics[width=0.95\linewidth]{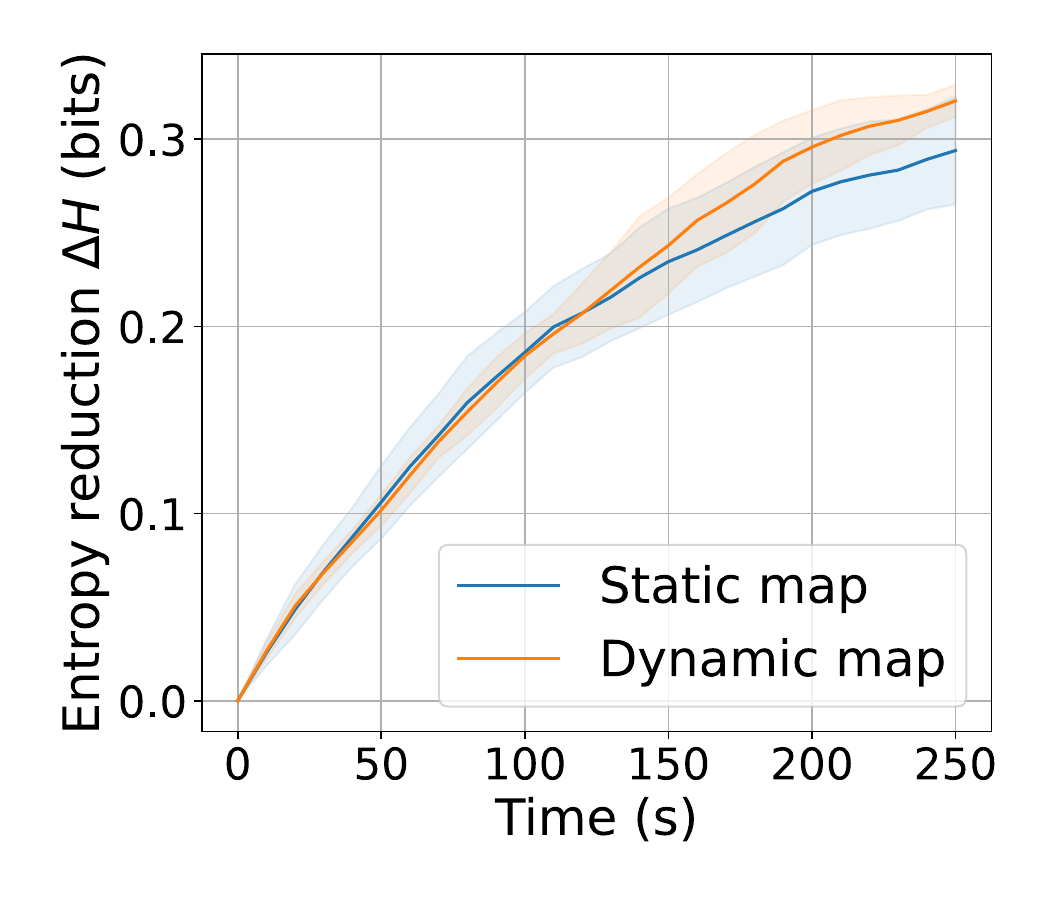}
    \end{subfigure}
    \begin{subfigure}{0.23\textwidth}
        \centering
        \includegraphics[width=0.95\linewidth]{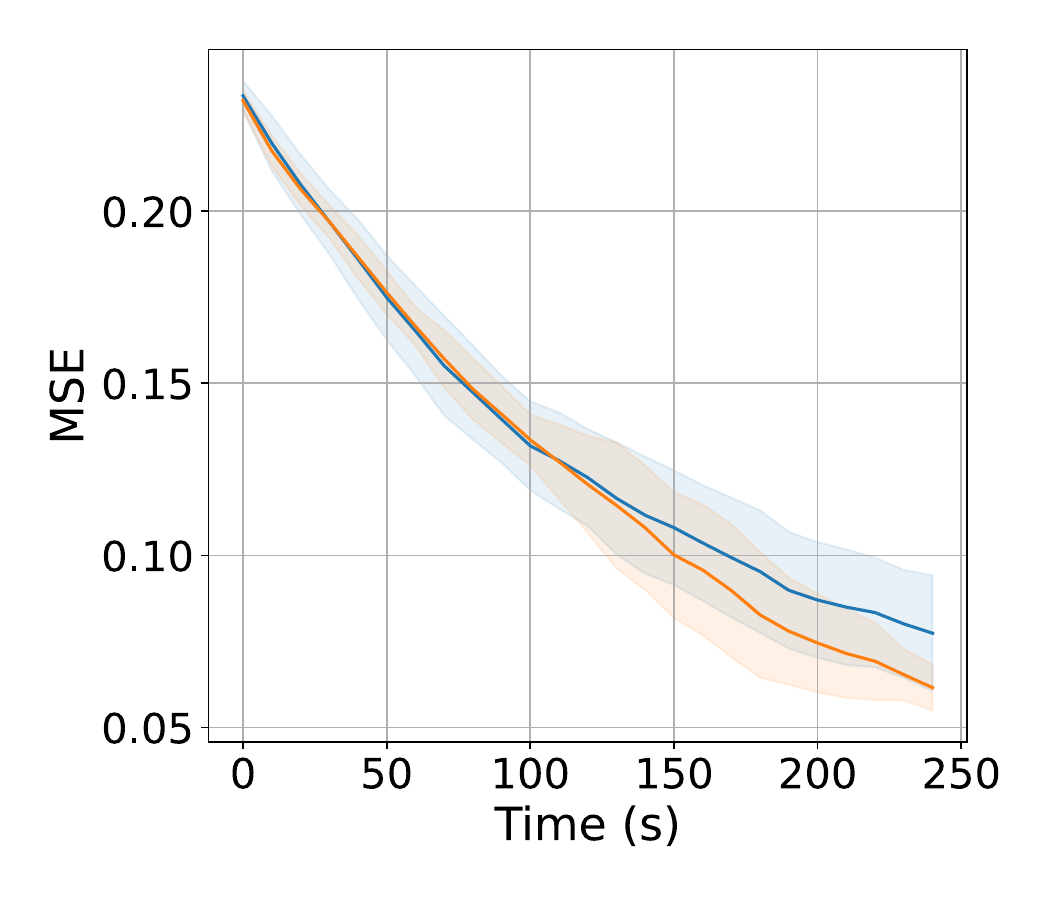}
    \end{subfigure}
    \caption{Entropy reduction $\Delta H$ and MSE for planning with the sampling-based planner, with and without the prediction step of the dynamic occupancy grid. With the prediction step, the mapping accuracy is observed to be better by $21.8\%$.}
    \label{fig:ablation2}
\end{figure}

To validate the advantages of our proposed prediction step of the dynamic occupancy grid mapping approach, we also perform simulation experiments without the prediction step of the mapping. Without the prediction step, the mapping is static and only relies on the sensor observations, as it does not account for any changes in the environment. We plot the resulting mapping accuracies and entropy for Monte Carlo simulations with and without the prediction step in \autoref{fig:ablation2}. 

The reduction in entropy at the end of the mission is similar, i.e., the trials including the prediction step in the dynamic occupancy grid mapping  $8.4 \%$ , as compared to mapping without the prediction step. However, the MSE values are $21.8\%$ lower with our dynamic mapping approach as compared to the mapping without the prediction step. This indicates that the planner still effectively reduces uncertainty in the map in either case, as observed from the entropy. However, the mapping accuracy is better with the prediction step included, as the static occupancy grid does not account for the targets' motion. This could also lead to multiple detections of a single target on the map. Such errors in the mapping also subsequently influence the decision-making of the robot, leading to suboptimal planning. 

\subsection{Comparison With Baselines}

\begin{table*}[ht]
\centering
\begin{tabular}{|l|cc|cc|cc|}
\hline
 & \multicolumn{2}{c|}{N = 8} 
 & \multicolumn{2}{c|}{N = 16} 
 & \multicolumn{2}{c|}{N = 32} \\
 & $ \Delta H$ & $\overline{N}$  
 & $\Delta H$ & $\overline{N}$ 
 & $\Delta H$ & $\overline{N}$ \\
\hline
Sampling-based
 & $0.315 \pm 0.014$ & $1.469 \pm 0.296$ 
 & $0.304 \pm 0.016$ & $2.539 \pm 0.329$  
 & $\mathbf{0.305 \pm 0.023}$ & $5.158 \pm 1.016$ \\
Receding horizon
 & $0.295 \pm 0.019$ & $\mathbf{1.690 \pm 0.142}$ 
 & $0.266 \pm 0.014$ & $\mathbf{3.015 \pm 0.268}$ 
 & $0.258 \pm 0.012$ & $\mathbf{6.089 \pm 1.014}$ \\
Greedy
 & $\mathbf{0.346 \pm 0.013}$ & $1.335 \pm 0.247$ 
 & $\mathbf{0.315 \pm 0.026}$ & $2.490 \pm 0.416$ 
 & $0.302 \pm 0.017$ & $4.244 \pm 0.651$ \\
Coverage
 & $0.239 \pm 0.005$  & $0.555 \pm 0.43$ 
 & $0.242 \pm 0.006$ & $1.389 \pm 0.904$ 
 & $0.236 \pm 0.003$ & $1.809 \pm 0.159$ \\
Random
 & $0.179 \pm 0.064$ & $0.706 \pm 0.49$ 
 & $0.171 \pm 0.059$ & $1.407 \pm 0.93$ 
 & $0.136 \pm 0.043$ & $2.383 \pm 1.460$ \\
\hline
\end{tabular}
\caption{Entropy, and mean detections $\overline{N}$ compared for the considered planning strategies for cases with various numbers of targets. The receding horizon planner is best at redetecting targets, followed by the sampling-based planner. Both these planning strategies benefit from longer lookahead predictions with the network. The greedy and sampling-based planners are both nearly as effective in reducing entropy.}
\label{tab:metrics_planners}
\end{table*}

We also evaluate our proposed sampling-based planner by comparing it against other planning strategies. It is executed in both a finite horizon and a receding horizon manner. In the receding horizon implementation, replanning is done after $50\%$ of the previous path is traversed. We choose the following baselines: (i) a greedy planner that greedily considers only a single waypoint to calculate utility \eqref{eq:reward}; (ii) coverage planning with lawnmower motion, traditionally used for monitoring \cite{galceran2013survey}; (iii) a random planner that selects waypoints randomly.
The coverage and random planners are non-adaptive strategies that do not take new information into account during the decision-making process. For the other planning strategies, the factor $w$ is set to $5$. Experiments are also performed for scenarios with a varying number of objects $N = 8, 16, 32$ to evaluate the ability of our approach to generalize to larger numbers of targets.

The mean entropy $H$ and mean detections $\overline{N}$ for the various planning strategies are shown in \autoref{tab:metrics_planners}. We observe that the non-adaptive baselines perform the worst in exploring the map and also in redetecting targets. The greedy planner reports very similar scores in terms of entropy as compared to the sampling-based planner, indicating it is equally effective at exploring the environment. However, the receding horizon planner is the best at tracking targets in all cases, followed by the sampling-based planner. This can be attributed to the sampling-based and receding horizon planners computing the utility by predicting sensor measurements along the path for the entire planning horizon, as opposed to the greedy planner that considers only a single waypoint rather than optimizing the utility over a path. 

\section{Field Tests}
\label{sec:field_exp}
To validate our framework in a real-world scenario, we perform field tests with an ASV for mapping freely drifting spherical buoy markers, as shown in \autoref{fig:experiment_1}. The ASV is overactuated with four thrusters in the X-configuration. It is equipped with a Cube Orange IMU and a Here4 GNSS for navigation. It has a Jetson Orin Nano, and software is implemented in ROS2. Trials are performed on a $25$\,m$\,\times\,25$\,m area for a mission time of $150$\,s. We use a ZED 2i camera with $72^\circ$ horizontal FOV. Detections up to a distance of $15$\,m are considered, as the camera gives reliable depth estimates up to $15$-$20$\,m. We use similar inverse sensor models for mapping as described in Sec. \ref{sec:sm}, with parameters scaled accordingly. The ASV's advance speed is set to $0.5$\,m/s, and the lookahead time for the planning is $25$\,s.

\begin{figure}[!htb]
    \centering
    \begin{subfigure}{0.23\textwidth}
        \centering
        \includegraphics[width=0.95\linewidth]{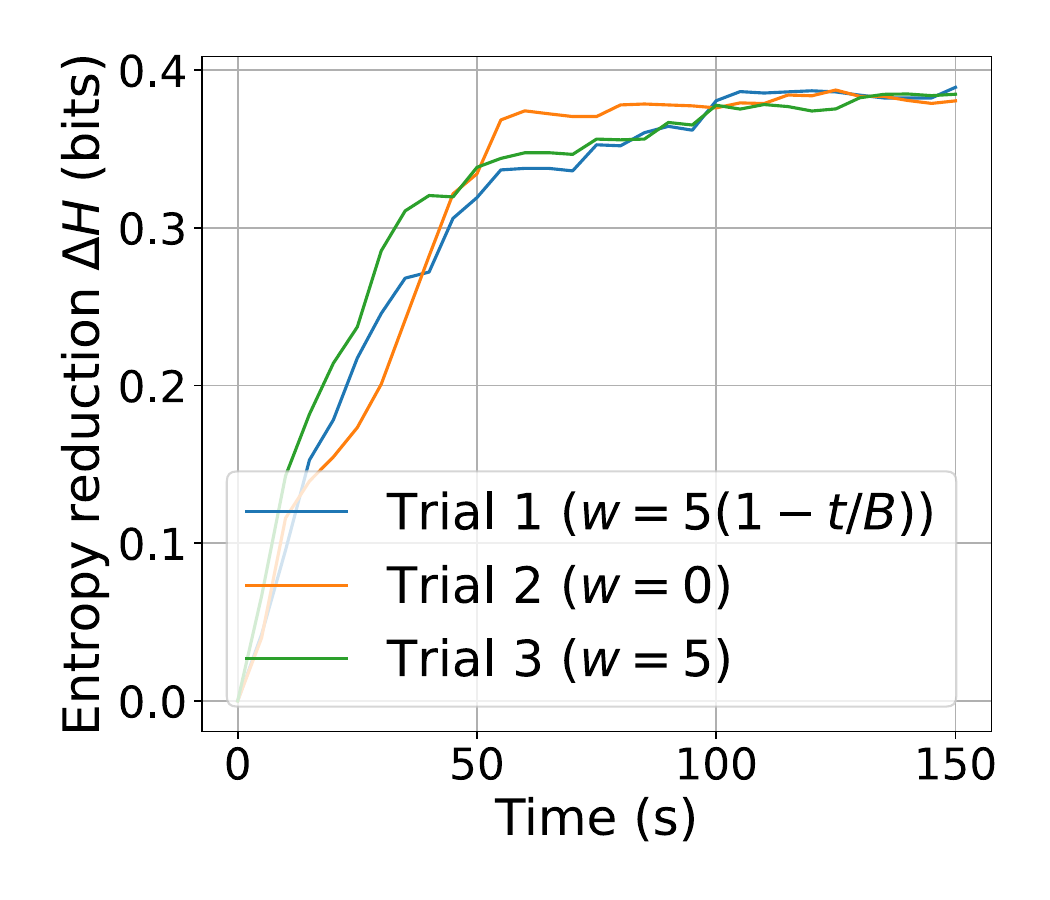}
    \end{subfigure}
    \begin{subfigure}{0.23\textwidth}
        \centering
        \includegraphics[width=0.95\linewidth]{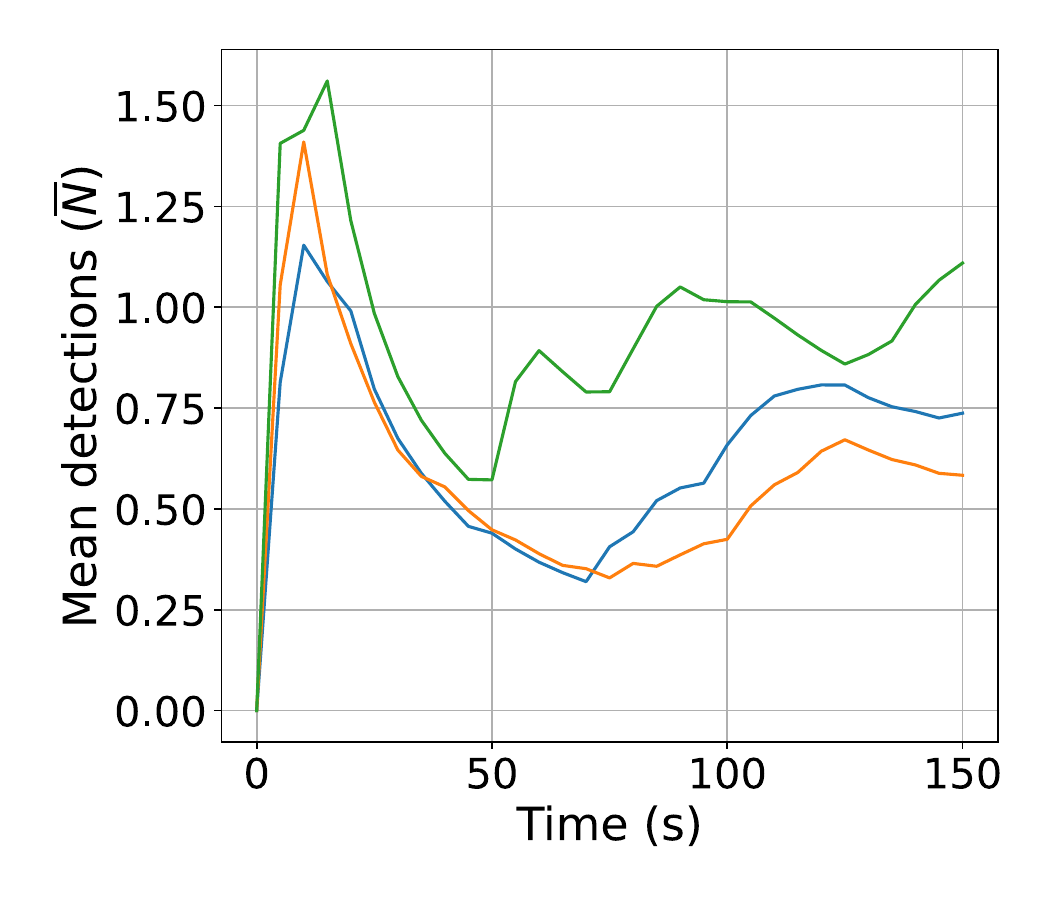}
    \end{subfigure}
    \caption{Evolution of $ \Delta H$ and mean detections $\overline{N}$ during field trials, where $w=5$ is most effective at redetecting targets.}
    \label{fig:exp_metrics}
\end{figure}

\begin{figure*}[!htbp]
    \centering
    \includegraphics[width=0.95\linewidth]{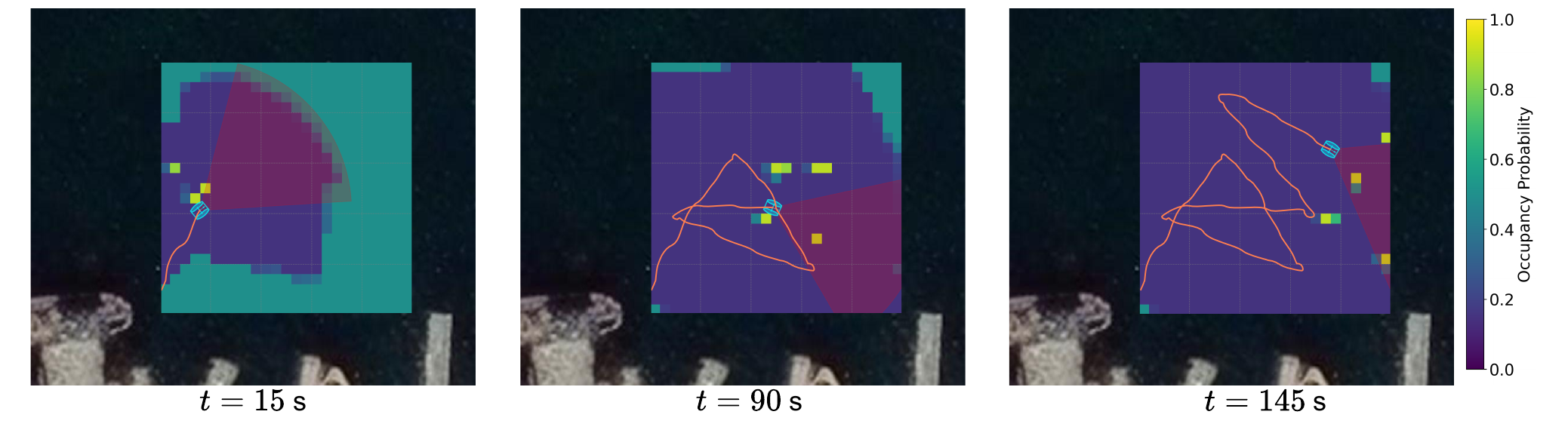}
    \caption{Occupancy grids plotted at different time instants from the field trial. The path followed is given in orange, the ASV pose and its FOV are given by the blue icon, and the red sector, respectively. At $t=15$ s, the ASV has mapped three buoys, and at $t=90$ s, it tracks the buoys again and has mapped all four buoys. The buoys are redetected again at $t=145$ s.}
    \label{fig:exp_plot}
\end{figure*}

During the experiments, moderate winds of $4$\,m/s were observed from West to East. We place four buoy markers near the left edge of the map at the start of each trial, and they are found to drift freely with the wind towards East due to the wind. We perform three trials with the sampling-based planner, each corresponding to $w=0, 5$, and $5(1-t/B)$, respectively, and discuss their results here. The resulting entropy and mean detections from the field trials are plotted in \autoref{fig:exp_metrics}. It is observed that the trial with $w=5$ is most effective at redetecting targets, as evident from several peaks in the graph for $\overline{N}$.
These results align with simulations showing that higher values of $w$ promote target reacquisition, further highlighting the usefulness of our proposed utility.
We also show the occupancy maps at various instants for this trial in \autoref{fig:exp_plot}. The ASV first detects three of the four targets at $t=15$\,s. We notice that at later intervals, $t=90$\,s and $145$\,s, the ASV tracks all four targets. 
An accompanying video of this experiment is also provided\href{https://youtu.be/nT_s3GbHHKU}{\footnotemark}.

\footnotetext{A video from the field trials with the ASV is available at \url{https://youtu.be/nT_s3GbHHKU}.}


The field experiments are limited in that we do not have access to an anemometer to measure the wind velocity at each instant. Therefore, we assume the wind speed to be constant during the trials for the prediction step in the mapping. From \autoref{fig:exp_plot}, we see that the ASV can still track and redetect targets. This can be attributed to our IPP approach, which takes into account uncertain target positions with the help of the spatiotemporal network, enabling informative paths for target re-detection despite possible inaccuracies in the prediction step of the mapping.

\section{Conclusions and Future Work}
\label{sec:conc}

In this study, we introduce an IPP framework to actively map moving targets in dynamic environments. We introduce a spatiotemporal prediction network to estimate uncertain target position distributions over time due to drift, and a new IPP utility for target tracking. We use a dynamic occupancy mapping approach to maintain an accurate map. Simulation results show that our proposed utility improves target tracking by redetecting targets more frequently, as compared to existing methods that use only entropy as the utility. Ablation studies further highlight the importance of our spatiotemporal network and our dynamic mapping method. Finally, field experiments with an ASV demonstrate the validity of our framework in real-world monitoring scenarios. 
While our approach can track targets efficiently in real-time, it relies on wind measurements. Therefore, estimating the drift of floating objects purely based on observations is an avenue for further work.
Another direction for future work is multi-robot spatiotemporal monitoring.

\section*{Acknowledgments}

The authors thank Joel Jose, Jon Urcelay, and Elisha Adiburo for assistance with the field experiments.

\bibliographystyle{IEEEtran}
\bibliography{references}

\end{document}